\documentclass{article}
\usepackage[a4paper, margin=1in]{geometry}
\usepackage{cite}
\usepackage{amsmath,amssymb,amsfonts}
\usepackage{graphicx}
\usepackage{textcomp}
\usepackage{float}
\usepackage{algorithm}
\usepackage{algpseudocode}
\usepackage{subcaption}
\usepackage{xcolor}
\usepackage{float}
\usepackage{babel,blindtext}
\usepackage{pdflscape}
\usepackage{array}
\usepackage{multirow}
\usepackage{enumitem}
\usepackage{textcmds}

\def\BibTeX{{\rm B\kern-.05em{\sc i\kern-.025em b}\kern-.08em
    T\kern-.1667em\lower.7ex\hbox{E}\kern-.125emX}}

\newlist{tabitemize}{itemize}{1}
\newlist{soloitemize}{itemize}{1}
\setlist[tabitemize,soloitemize]{
  nosep, nolistsep,
  topsep=6pt,
  align=left,
  left=0pt,
  label=$\bullet$,
}

\usepackage{bm}
\makeatletter
\AtBeginDocument{\DeclareMathVersion{bold}
\SetSymbolFont{operators}{bold}{T1}{times}{b}{n}
\SetMathAlphabet{\mathrm}{bold}{T1}{times}{b}{n}
\SetMathAlphabet{\mathit}{bold}{T1}{times}{b}{it}
\SetMathAlphabet{\mathbf}{bold}{T1}{times}{b}{n}
\SetMathAlphabet{\mathtt}{bold}{OT1}{pcr}{b}{n}
\SetSymbolFont{symbols}{bold}{OMS}{cmsy}{b}{n}
\renewcommand\boldmath{\@nomath\boldmath\mathversion{bold}}}
\makeatother

\def\BibTeX{{\rm B\kern-.05em{\sc i\kern-.025em b}\kern-.08em
    T\kern-.1667em\lower.7ex\hbox{E}\kern-.125emX}}

\begin{document}

\title{Affordance Labeling and Exploration: A Manifold-Based Approach}
\author{İsmail ÖZÇİL$^{1}$ , A. Buğra KOKU$^{1,2}$
\\
{\footnotesize [1]{Department of Mechanical Engineering, METU, Ankara, 06800, Türkiye(e-mail: iozcil@metu.edu.tr)}}\\
{\footnotesize [2]{Center of Robotics and Artificial Intelligence, METU, Ankara, 06800, Türkiye (e-mail: kbugra@metu.edu.tr)}}}
\maketitle
\begin{abstract}
The advancement in computing power has significantly reduced the training times for deep learning, fostering the rapid development of networks designed for object recognition. However, the exploration of object utility, which is the affordance of the object, as opposed to object recognition, has received comparatively less attention. This work focuses on the problem of exploration of object affordances using existing networks trained on the object classification dataset. While pre-trained networks have proven to be instrumental in transfer learning for classification tasks, this work diverges from conventional object classification methods. Instead, it employs pre-trained networks to discern affordance labels without the need for specialized layers, abstaining from modifying the final layers through the addition of classification layers. To facilitate the determination of affordance labels without such modifications, two approaches, i.e. subspace clustering and manifold curvature methods are tested. These methods offer a distinct perspective on affordance label recognition. Especially, manifold curvature method has been successfully tested with nine distinct pre-trained networks, each achieving an accuracy exceeding 95\%. Moreover, it is observed that manifold curvature and subspace clustering methods explore affordance labels that are not marked in the ground truth, but object affords in various cases.
\end{abstract}


\section{Introduction}
\label{sec:introduction}
Object recognition involves identifying objects using sensor data, typically captured through images or videos from a camera. For a robot to interact effectively with its environment, it needs to recognize and detect objects in its sensor space and then use these recognized and detected objects in its parameter space when required.

Deep neural networks, such as LeNet~\cite{LeCun1988}, AlexNet~\cite{krizhevsky2012imagenet}, VGG \cite{Simonyan2015}, Inception and GoogLeNet \cite{szegedy2015going}, and ResNet \cite{he2016deep}, have enabled object classification in various domains. However, compared to classification, there has been relatively less focus on defining object characteristics and understanding their usage. Interacting with the environment requires not only identifying objects correctly but also defining their properties and estimating possible usages based on these properties.

The concept of "affordance", introduced by Gibson\cite{gibson1966senses}, is essential in this context. Affordance refers to a specific combination of an object's substance and the surfaces about an animal. Later, Gibson modified this concept to relate the perception of an object to its potential actions\cite{gibson}. For a robot, estimating the affordances of an object is crucial for effective interaction with people, its environment, and other robots. By estimating the affordances of objects, robots can understand how to use them effectively for various tasks. This highlights the importance of robots not just recognizing objects, but also perceiving their potential uses in different scenarios.
\subsection{Objective}
The objective of this work is to explore methodologies enabling a robot to acquire both the affordance features and labels associated with objects in its operational environment. While studies mentioned in the previous part are object classification networks, this study focuses on the domain of affordance. Affordance presents a distinct challenge from object classification, as various objects can facilitate similar actions, and a single object often possesses multiple affordances. In this context, "affordance labeling" pertains to characterizing the affordances of objects in a given scenario. For example, it involves labeling actions such as grasping and liquid containment for different images containing different objects that are graspable or capable of containing liquids. Moreover, affordance labeling of an image containing an object is also a challenge as the number of affordance labels is not fixed, even in the same object groups. It should be noted that, affordance of an object perceived by humans is subjective. People might discover affordance of an object after some interaction with it. Hence, ground truth labels in relevant datasets do not necessarily contain labels for everything an object can afford.  

Consequently, one-hot encoding approaches are ill-suited for solving affordance labeling tasks. Furthermore, because objects frequently serve multiple affordances, the datasets associated with objects that support specific affordances may intersect or overlap. This overlap results in shared data points or regions within the corresponding affordance label groups such as the same area of a bottle may afford `grasp' and `roll' affordances. Hence, making it hard to label the corresponding area. Moreover, there may be affordances of the given objects at first sight. Exploration of the hidden affordances is also important. Thus, the core challenge at hand involves defining the affordance labels for given objects based on their images taking into account that one object may afford multiple affordance categories. 

\subsection{Related Work}
Uğur et al.~\cite{ugur2015staged} presented a framework for a robot manipulator to understand human infant sensorimotor skills. Their framework consists of two stages: developing movement capability and behavior primitives, constructing affordance predictors based on cause-effect relations, and receiving guidance from a tutor to imitate actions. They explored grasping, releasing, touching, and no-touching primitives but plan to add attention ability and relational affordances in the future. Modayil and Kuipers\cite{modayil2008initial} also explored unsupervised learning of affordances using sensorimotor data during object interactions. Koppula et al.\cite{Koppula2013} modeled human activities and object affordances in RGB-D videos using Markov random fields. Hartson\cite{Hartson2003} expanded the concept of affordance, defining four categories: cognitive, physical, sensory, and functional affordance. There are also studies on affordances for robot manipulation tasks. Iriondo et al.\cite{iriondo2021affordance} estimated grasping points using graph convolutional networks and point cloud data.  Wu and Chirikjian\cite{Wu2020} investigated strategies for pouring objects using an RGB-D camera and simulation to determine pourability. Ardón et al.\cite{ardon2021affordance} focused on robot-to-human object handovers, considering human comfort and object usability. There are also datasets that can be used in affordance research. Khalifa and Shah\cite{khalifalarge} introduced a large-scale multi-view RGBD dataset containing object images with affordance labels. Chuang et al.\cite{cychuang2017learning} constructed a dataset with people interacting with objects and offered a new method to determine object affordances and human-object interactions. Various approaches have been proposed for learning affordances from visual data. Wenkai et al.\cite{wenkai} present a method for 6-DoF grasp detection that leverages both implicit neural representation and visual affordance estimation, while Li et al.\cite{liLocate} introduce LOCATE, a framework that focuses on localizing and transferring object parts for weakly supervised affordance grounding. Several studies focused on methods to infer 3D affordances from 2D information. Yang et al.\cite{yangGrounding} propose a method to ground 3D object affordances from 2D images depicting human interactions, while Yang et al.\cite{yang2023lemon} present an approach for learning 3D human-object interaction relations from 2D images. Affordance learning faces challenges such as handling open vocabularies of unseen affordances and limited data availability. Vo et al.\cite{voOpenVocab} address the open vocabulary challenge by proposing a method for affordance detection using knowledge distillation and text-point correlation. Li et al.\cite{li2023oneshot} tackle the data limitation issue by introducing a one-shot learning approach with foundation models, demonstrating effective affordance learning with minimal training data. Ragusa et al.\cite{ragusa} focus on developing efficient models for wearable robots, proposing tiny networks for affordance segmentation on resource-constrained devices. Chen and Chenyi \cite{chen2015deepdriving} proposed a method to map an input image to driving affordances using ConvNet, providing a more controllable approach to autonomous driving compared to the direct mapping of commands. Andries et al. \cite{Andries2020} designed 3D objects based on required functionalities by training a neural network to relate function to form using a dataset of affordance-labeled objects. Ruiz and Mayol-Cuevas \cite{ruiz2020geometric} used interaction tensors to estimate affordance possibilities for various actions, but predicting affordances for flexible objects was challenging. Moldovan et al. \cite{moldovan2012learning} learned affordances for multiple interacting objects using statistical relational learning. Thermos et al. \cite{thermos2020deep} obtained pixel-level affordances of images by training an auto-encoder network with human-object interaction videos. Uğur et al. \cite{ugur2009affordance}\cite{Ugur2011}\cite{Ugur2007} utilized robotic hand actions and observations to learn object affordances, incorporating behavioral parameters and recognizing traversable objects in a room. Bozeat and Ralph \cite{bozeat2002objects} studied the effect of prior knowledge on object usage by patients and found that prior knowledge has a significant impact on object usage. Federico and Brandimonte\cite{Federico2020} investigated tool usage characteristics of people and found support for reasoning-based theories of human tool use. Şahin et al.\cite{csahin2007afford} discussed the affordance concept and proposed a new formalization for autonomous robots. Xu et al. \cite{deepaff} introduced a method for predicting future affordance states following an action applied to an environment. Ransikarbum et al. \cite{highway} developed a model for driver decision-making using road affordances, aimed at enhancing autonomous driving systems. Hassanin et al. \cite{anewloc} made improvements to Mask-RCNN to address scaling issues in determining object part affordances at various scales. Myers et al. \cite{affdetecttool} utilized RGBD images to extract feature vectors and made affordance decisions through Support Vector Machine and Structured Random Forest methods. Pandey and Alami \cite{affgraph} proposed a framework to enhance human-robot interaction by leveraging affordances to perform contextually appropriate and relevant actions. Nguyen et al. \cite{detectingaff} employed Convolutional Neural Networks (CNNs) to detect object affordances. Ragusa et al. \cite{hardwareaff} presented a method for real-time affordance detection on resource-constrained systems. Thermos et al. \cite{joaff} developed a model to detect affordances of object parts in RGBD videos. These studies compose an understanding of the concept of affordances, problems of applying the concept of affordances into the field of robotics and methods to solve these problems.

\subsection{Contributions of the Study}
Within this work, we make the following contributions:
\begin{itemize}
    \item Demonstrating the Applicability of Pre-Trained Classifiers on ImageNet\cite{Imagenet} for Affordance Detection: We present the practicality and effectiveness of employing pre-trained classifiers in the task of affordance detection without transfer learning.
    \item Labeling affordances based on feature vectors: We employ vector-based labeling methods to omit the last fully connected layer. Instead of adding and training a new fully connected layer to fit the requirements of labeling, two different decision methods based on vector properties are proposed.
    \item Utilizing Manifold Angles for Affordance Labeling: We introduce a novel approach that leverages manifold angles to make informed affordance labeling decisions without the necessity of introducing additional layers to existing pre-trained architectures. 
    \item Exploration of affordances: Our proposed labeling methods exhibit the capability to identify potential affordances in certain cases not annotated in the ground truth image. 
\end{itemize}
These contributions collectively enhance our understanding of affordance detection and exploration. An indirect contribution of this work is on the problem of multi-labelled classification problems, where there is a many to many mapping between inputs and the labels.
\section{Method}
In this study, we make use of an affordance dataset containing RGBD images of various objects, accompanied by segmentation labels that specify affordance labels for each pixel. While most related works discussed in the preceding section relied on depth images to estimate object affordances, our approach takes a different path by omitting depth image data from our affordance labeling method.

There exist two types of affordance labeling methods. The first method requires segmenting and labeling individual pixels within an image based on their respective affordances. The second method, which is the focus of our investigation, does not make pixel or segment labeling and instead assigns predicted affordances directly to the entire image. To validate our proposed methodologies, we specifically concentrate on image-level labeling without any segmentation.

In addition to the affordance dataset they provided, Khalifa and Shah\cite{khalifalarge} presented various approaches, some of which involved employing pre-trained networks for feature extraction. Typically, these methods entailed appending and training a new fully connected layer to generate estimations tailored to the specific dataset. In contrast, our present study differs from this practice by only using pre-trained networks for the obtaining of feature vectors. Unlike methodologies that incorporate and train new fully connected layers, our approach avoids directly obtaining results from the tailored fully layer. Instead, we harness the feature vector derived from the last layer before the fully connected layer of the original network as shown in the Figure \ref{fig:basecnn} to formulate estimation criteria through two distinct methods. The rationale behind opting for feature vectors lies in their capacity to afford flexibility in the number of affordance decisions, allowing for easier integration of new affordance classes into an existing decision structure. Furthermore, the utilization of vectors facilitates the exploration of novel affordances associated with a given object, predicated on the "closeness" to other object vectors with different affordance labels. Additionally, adopting vectors for decision-making allows us to explore manifolds where feature vectors of various affordance classes are situated.

\subsection{RGBD Affordance Dataset}
We employ the dataset named `A large scale multi-view RGBD visual affordance learning dataset\cite{khalifalarge} which provides object affordance annotations at the pixel level to test our proposed methods. The dataset is composed of 23605 scenes, with every scene having 1 RGB, 1 depth, and 1 label image from 37 object categories annotated with 15 affordance categories, which are `grasp', `wrap grasp', `containment', `liquid-containment', `openable', `dry', `tip-push', `display', `illumination',  `cut', `pourable', `rollable', `absorb', `grip' and `stapling'. 

The `grasp' affordance refers to parts of objects that can be grasped by the user, while `wrap grasp' labeled to objects that can be held by wrapping fingers around them. `Containment' is associated with bowl-like objects capable of containing other materials, and `liquid-containment' denotes bottle-like parts of objects that can hold liquids. For consistency, `contain' and `liquid contain' are used instead of `containment' and `liquid-containment', respectively. The `openable' affordance refers to cap-like or lid-like parts of objects that can be opened by the user. For consistency, we use `open' instead of `openable' throughout this study. The `dry' affordance identifies parts of objects used to dry other surfaces or objects.

`Tip-push' describes button-like parts of objects that can be pressed, while `display' identifies objects equipped with displays. `Illumination' refers to objects that emit light. For consistency, `illuminate' is used instead of `illumination'. The `cut' affordance applies to objects capable of cutting other objects, and `pourable' describes objects designed to pour liquids. We use `pour' instead of `pourable' for consistency. `Rollable' refers to objects that can be rolled. We use `roll' instead of `rollable'. The `absorb' affordance identifies objects with porous structures that can absorb liquids, and `grip' pertains to objects with parts designed for a secure grip. Finally, `stapling' refers to staples. For consistency, we use `stable' instead of `stapling'.

These affordance labels from the dataset and their corresponding terms used in this study are summarized in Table \ref{tab:expln}.
\begin{table}[H]
\caption{\textbf{Dataset and Current Study Affordance Labels\cite{khalifalarge}}}
\begin{center}
\begin{tabular}{|c|c|}
\hline
Label Used in Dataset  & Label Used in this Study\\
\hline
grasp & grasp\\
wrap-grasp & wrap-grasp\\
contain & contain\\
liquid contain & liquid contain\\
openable & open\\
dry & dry\\
tip-push & tip-push\\
display & display\\
illumination & illuminate\\
cut & cut\\
pourable & pour\\
rollable & roll\\
absorb & absorb\\
grip & grip\\
stapling & staple\\
\hline
\end{tabular}
\label{tab:expln}
\end{center}
\end{table}
The dataset is composed of scene images in which a single object is presented. These objects are pixel-level annotated with corresponding affordance labels in their label images. Parts of the objects are labeled with suitable affordance labels. Due to the pixel-level annotation, each region can be labeled with only one affordance label. However, a single part or region of an object may afford multiple affordances. This opens the possibility to `explore' new affordances if they are available in the dataset.

\begin{figure}[H]
\begin{tabular}{ccc}
  \includegraphics[width=50mm]{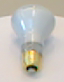} & 
  \includegraphics[width=50mm]{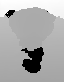}&
  \includegraphics[width=50mm]{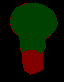} \\
(a) RGB Image & (b) Depth Image & (c) Affordance Label\\
 & & Image \\
\includegraphics[width=50mm]{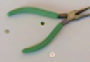} & 
  \includegraphics[width=50mm]{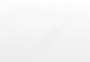}&
  \includegraphics[width=50mm]{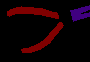} \\
(a) RGB Image & (b) Depth Image & (c) Affordance Label\\
 & & Image \\
\end{tabular}
\caption{Dataset examples of the objects `lightbulb' and `pliers'\cite{khalifalarge}}
\label{fig:dataset_examples}
\end{figure}

In addition to the scenes with single objects, the dataset also contains 35 cluttered/complex test scenes with different objects and multiple affordances. Figure \ref{fig:dataset_examples} illustrates some examples from the dataset. Although they are not used in any part of the study, depth images are scaled to be represented by a grayscale image to be shown as dataset example.
Notably, the image dimensions across the dataset vary; different scenes may exhibit distinct width and height values. Occasionally, some images possess dimensions incompatible with pre-trained networks. To address this challenge, we implemented zero padding for images with disparate heights and widths to render them rectangular. Subsequently, we employed upscaling or downscaling techniques based on the adjusted image dimensions to standardize the size of every image within the dataset.

\subsection{Pre-trained Neural Network}
\begin{figure}[H]
	\begin{center}
		\includegraphics[scale=0.2]{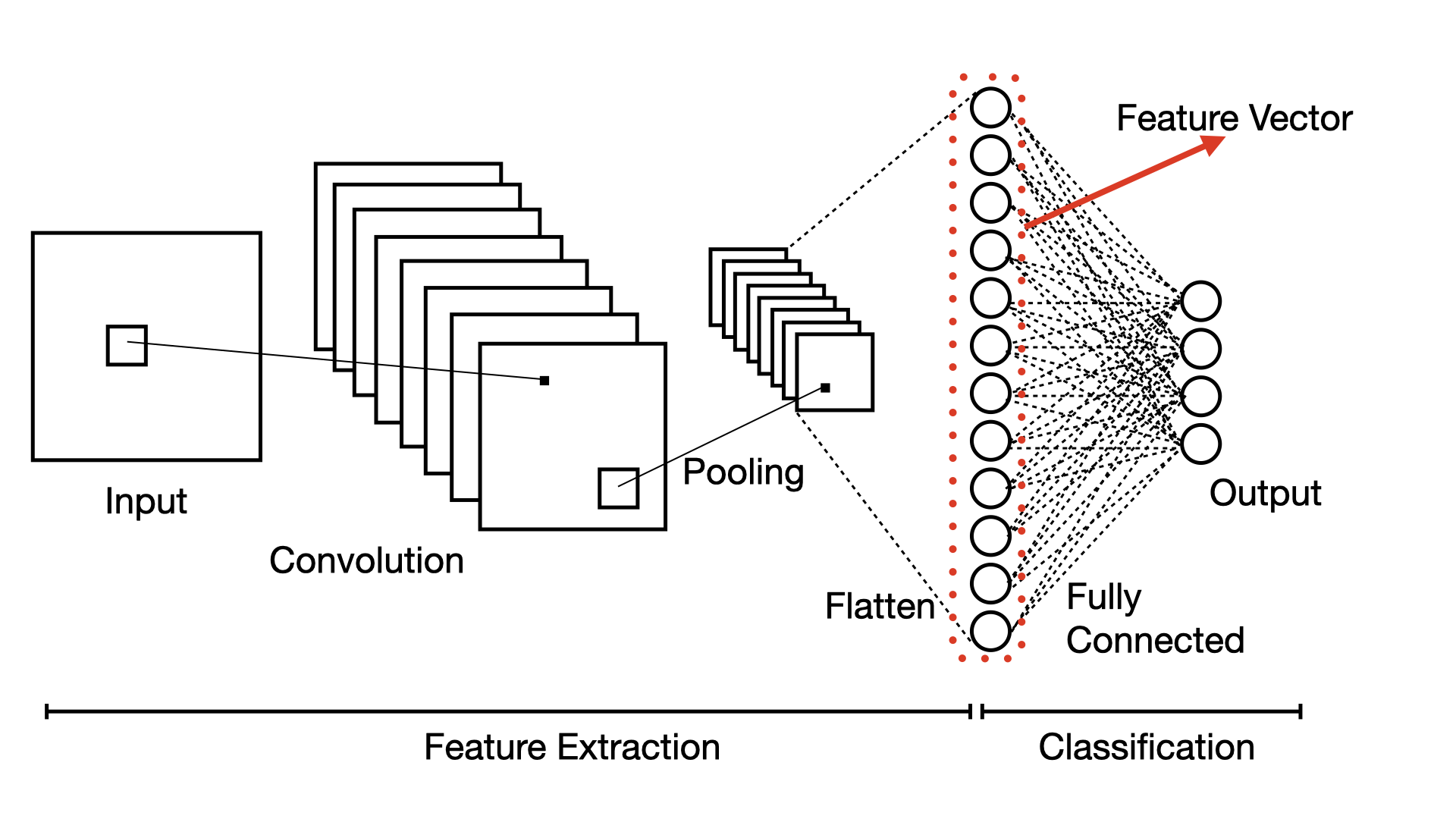}
		\caption{A simple CNN with feature vector extraction}
		\label{fig:basecnn} 
	\end{center}
\end{figure}

As pre-trained networks, ResNet-18, ResNet-50, ResNet-101, ResNet-152\cite{resnet18}, ResNext-101\cite{resnext}, RegNetY\cite{regnety}, Efficient-NetV2\cite{efficientnetv2}, ViT-L/16, and ViT-B/16\cite{vit} trained on the ImageNeT\cite{Imagenet} dataset have been selected to generate feature vectors from the previously explained affordance dataset. ImageNet\cite{Imagenet} is a dataset used for object recognition tasks. The final fully connected layers of these pre-trained networks are removed to obtain feature vectors as outputs as shown in Figure \ref{fig:basecnn}. Due to the distinctive architectures of these networks, the dimensions of their respective feature vectors vary. As explained in the preceding section, the dataset utilized contains RGB images paired with depth images and corresponding affordance annotations at the pixel level. In this context, only the RGB images of objects serve as inputs for the modified networks to extract feature vectors. The depth images are disregarded throughout this study and are not used at any stage. While pixel-level object usability labels are not directly utilized, they serve as image-level labels for our experiments.  Each object image is annotated with a combination of object usability labels derived from its pixel annotations, rather than using the pixel-level annotations themselves.  
Multiple networks were employed to assess the labeling performance of feature vector outputs generated by each network. The utilization of multiple networks facilitates not only the evaluation of individual labeling performances of the pre-trained networks but also the comparison of the robustness of labeling methods based on the proposed feature vectors. We target developing a method that has good performance with every one of the previously mentioned networks. This approach enables a comprehensive analysis of the effectiveness and stability of the labeling techniques under consideration.

For deep learning applications, a dataset generally is divided into two parts; training set and validation or test set. Since the pre-trained neural network is planned to be used as it is without its last fully connected layer, there is no need for training for this case. Instead, decisions are made according to the properties of the data which are extracted from a certain part of the dataset. Similarly, for training, the dataset is divided into learning and validation parts. "Learning" term is used here since there is no layer by layer training in our methods where learning involves analysis of feature vectors that are obtained from the part of the dataset that is set apart for this purpose. Out of a total of 23,605 scenarios from the affordance dataset, 18,000 images are allocated for the learning, while 5,605 images are reserved for validation purposes. The feature vectors corresponding to the learning data are then analyzed and categorized based on the affordance labels associated with each dataset entry. Algorithm \ref{alg:data_sep} shows the process of generating affordance groups from the affordance labels using the learning set. It's important to acknowledge that a single feature vector may be linked to multiple affordance categories, reflecting the potential for an object to serve multiple purposes and, consequently, carry multiple labels. 

\begin{algorithm}[H]
\caption{Dataset Feature Extraction}\label{alg:data_sep}
\textbf{Input:} $r$: RGB images,
\\$label\_img$: Affordance segmented label images, \\$affordancess$: List of all affordance labels\\
\textbf{Output:} $C_{learning}$, $C_{validation}$ :List of learning and validation feature vector sets.
\begin{algorithmic}
\Require $f:$ Pre-Trained Network as Feature Extractor. 
\State Separate $r$ to $r_{learning}$ and $r_{validation}$
\State Initialize $C_{learning}$
\For{$i$ in $affordances$}
    \State Initialize $C_{i,learning}$.
    \For{$scene\_image$ in $r_{learning}$}
        \If{$label\_img(scene\_image)$ has $i$}
            \State Append $f(scene\_image)$ to $C_{i, learning}$
        \EndIf
    \EndFor
    \State Append $C_{i, learning}$ to $C_{learning}$
\EndFor
\State Initialize $C_{validation}$
\For{$scene\_image$ in $r_{validation}$}
    \State $j$ $\gets$  $f(scene\_image)$
    \State Append $j$ to $C_{validation}$
\EndFor
\State $\mathbf{return}$ $C_{learning}$, $C_{validation}$

\end{algorithmic}
\end{algorithm}

\subsection{Subspace Projection Approach}
The first method for labeling affordances using feature vectors, obtained from the aforementioned pre-trained networks, serves as a baseline reference. This method operates under the assumption that feature vectors corresponding to the same affordance are situated within an affine space. Hence, the learning part of the dataset is based on the extraction of subspace base vectors for each affordance group. In the learning group, each feature vector corresponding to images are labeled with its ground truth affordance group. Then, using all the feature vectors for a particular affordance group from the learning set, the basis vectors of the affordance group are found using Singular Value Decomposition (SVD). Here, each feature vector of the same affordance group are stacked to form the columns of a matrix which is then decomposed using SVD.

To label the images using subspace projection, we first identify the basis vectors for the subspaces corresponding to each affordance label. Initially, for each affordace group $i$, we stack the learning set feature vectors of that group into a matrix. Then, the obtained matrix is decomposed using SVD to get basis vectors of the affordance group $i$ as shown in the equations~\eqref{stackeqn} and~\eqref{svdeqn}. We assume that each group occupies a space with a predetermined dimension denoted as `$d$'. From the resulting U matrix obtained via SVD, we select the first `$d$' column vectors. These vectors are designated as the basis for the respective affordance space.
\begin{equation}
\mathbf{M_i} = horizontalStack(\mathbf{C_{i,learn}})
\label{stackeqn}
\end{equation}
\begin{equation}
\mathbf{U_i} \mathbf{\Sigma_i} \mathbf{V^T_i} = \mathbf{M_i}
\label{svdeqn}
\end{equation}

\begin{figure}[H]
    \centering   \includegraphics[scale =0.8]{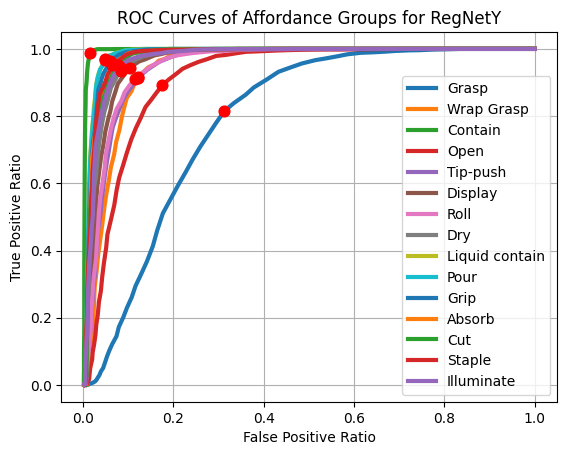}
    \caption{ROC curves of pre-trained RegNetY output}
    \label{fig:roc-regnety}
\end{figure}

 Then, the most significant $d$ number of basis vectors of the $\mathbf{M_i}$ are selected as $\mathbf{U}_i = \begin{bmatrix} u_1 & u_2 & \hdots & u_d\end{bmatrix}$. Obtained $\mathbf{U}_i$  is a matrix whose columns form an orthonormal basis for the $d$ dimensional subspace of the $\mathbf{M}_i$. Projection matrix $\mathbf{P}_i$ for $d$ dimensional subspace of the affordance group $\mathbf{M}_i$ is then given in equation~\eqref{projectionmatrix} since columns of $\mathbf{U}_i$ are already orthonormal. 
\begin{equation}
\mathbf{P}_i = \mathbf{U}_i \mathbf{U}_i^T
\label{projectionmatrix}
\end{equation}
The calculation process of the projection matrices is shown in the algorithm \ref{alg3:pm}.
\begin{algorithm}[H]
\caption{Projection Matrix Calculation}\label{alg3:pm}
\textbf{Input:} $C_{learning}$: List of feature vectors of the learning set of all affordance groups\\
\textbf{Output:} $P$: List of projection matrices for all affordance groups 
\begin{algorithmic}
\State Initialize $\mathbf{P}$
\For{$C_{i,learning}$ in $C_{learning}$ with affordance group $i$}
    \State $\mathbf{M_i} \gets horizontalStack(C_{i,learning})$  
    \State $\mathbf{U_i}, \mathbf{\Sigma_i}, \mathbf{V^T_i} \gets SVD(\mathbf{M_i})$
    \State $\mathbf{P}_i \gets \mathbf{U}_i \mathbf{U}_i^T$
    \State Append $\mathbf{P}_i$ to $\mathbf{P}$ 
\EndFor

\State $\mathbf{return}$ $\mathbf{P}$
\end{algorithmic}
\end{algorithm}
To determine the appropriate affordance label(s) for an image from the validation set, the image's feature vector is projected onto each of the previously determined pre-defined subspaces. Subsequently, the ratio of the $l_2$ norm of the projected vector to that of the original vector is computed. Let $j$ be a feature vector from the validation set of the dataset, $C_{validation}$. Then, the $l_2$ norm ratio of the projected feature vector of the $j$ onto $d$ dimensional subspace of the affordance group $\mathbf{M}_i$ to original vector $j$  shown in the equation~\eqref{l2ratio}.
\begin{equation}
    d_{i,j} = \frac{|Proj_i(j)|_2}{|j|_2}
    \label{l2ratio}
\end{equation}
If this ratio surpasses the threshold value, the corresponding affordance label is assigned to the image. To ascertain an appropriate threshold value, a systematic process is followed as shown in the Algorithm \ref{alg3:thres}. Firstly, for each affordance group, all instances of the learning part of the feature vectors, $C_{learning}$, are investigated. For each affordance $i$, projection ratios of the feature vectors labeled with affordance $i$ and non-labeled feature vectors with the affordance $i$ are stored. Then, using True Positive (TP), False Negative (FN), False Positive (FP), and True Negative (TN) values True Positive Ratio (TPR) and False Positive Ratios (FPR) are calculated for each threshold value ranging from $0$ to $1$ in an incremental fashion. TPR and FPR calculations are shown in ~\eqref{eq:tpr} and ~\eqref{eq:fpr}, respectively. The resulting TPR vs FPR values against increasing threshold values are given in Figure \ref{fig:roc-regnety}. 
\begin{equation}
TPR = \frac{TP}{TP + FN}
\label{eq:tpr}
\end{equation}

\begin{equation}
FPR = \frac{FP}{FP + TN}
\label{eq:fpr}
\end{equation}
After getting the TPR and FPR values for each threshold value for each affordance group, the threshold score denoted by $ts$ for threshold values ranging from $0$ to $1$ for each affordance group separately are determined according to equation~\eqref{eq:opt}:
\begin{equation}
ts = \sqrt{(1-TPR)^2+FPR^2}
\label{eq:opt}
\end{equation}
Presented $ts$ value in equation~\eqref{eq:opt} shows how close a particular point to the $TPR =1, \: FPR=0$ point, which is a represents $100\%$ labeling performance. Hence, this closeness value, $ts$ is used as a labeling performance metric. After the calculation of $ts$ values for each threshold value, the threshold value with the minimum $ts$ is chosen as the threshold value for the affordance group. Figure \ref{fig:roc-regnety} shows the optimal threshold TPR and FPR values for each affordance group with red dots on the graphs. As a result, the threshold values (shown as red dots  in Figure \ref{fig:roc-regnety}) are selected for each affordance group. Calculation of threshold values for each affordance group is presented in algorithm \ref{alg3:thres}. 
\begin{algorithm}[H]
\caption{Threshold Value Determination}\label{alg3:thres}
\textbf{Input:} $C_{learning}$: List of feature vectors of the learning set of affordance groups\\
$P$: List of projection matrices of affordance groups\\
\textbf{Output:} $th$: List of threshold values for each affordance group. 
\begin{algorithmic}
\State Initialize $th$
\For{$i$: affordance group in $affordances$}
    \State Initialize $l_i$: list of labeled projection ratios
    \State Initialize $nl_i$: list of unlabeled projection ratios
    \For{$\mathbf{j}$: feature vector in $C_{learning}$}
        \State $d_{i,j} \gets {norm(\mathbf{P}_i \mathbf{j})}/{norm(\mathbf{j})}$
        \If{$\mathbf{j}$ has label $i$}
            \State Append $d_{i,j}$ to $l_i$
        \Else
            \State Append $d_{i,j}$ to $nl_i$
        \EndIf
    \EndFor
    \State Initialize $th_i$ 
    \State Initialize $ts_i$
    \For{$thresh$ in $range(0,1)$}
        \State Calculate $ts_i$ 
        \If{$ts_i < previous\: ts_i$}
            \State $th_i = thresh$ 
        \EndIf
    \EndFor
    \State Append $th_i$ to $th$
\EndFor 

\State $\mathbf{return}$ $th$
\end{algorithmic}
\end{algorithm}
Up to this point, only the learning subset of the dataset has been employed to compute the basis vectors and threshold values for each distinct affordance group. The next phase involves the labeling of the validation data. The process of the affordance labeling of a given image by Subspace Projection Method (SPM) is given by the algorithm \ref{alg2:subsp}. 
\begin{algorithm}[H]
\caption{Affordance Labeling via SPM}\label{alg2:subsp}
\textbf{Input:} $C_{validation}$: List of feature vectors of the validation set of affordances ,\\ $th$: List of threshold values for Affordance groups,\\
$P$: List of affordance subspace projection matrices\\
\textbf{Output:} $aff$: List of indices of objects labeled with corresponding affordance. 
\begin{algorithmic}
\State Initialize $aff$
\State $\mathbf{M}_{validation} \gets horizontalStack(C_{validation})$ 
\For{$\mathbf{P}_i$:Projection matrix, $th_i$:threshold value of affordance $i$ in $P$ and $th$} 
\State $projections \gets \mathbf{P}_i \mathbf{M}_{validation}$
\State Append $args(projections>th_i)$ to $aff$
\EndFor
\State $\mathbf{return}$ $aff$
\end{algorithmic}
\end{algorithm}

As previously indicated, a subset of 5605 RGB images from the dataset has been allocated for validation. The performance of the SPM is computed by calculating TPR and FPR values for each validation scene. Then, overall TPR and FPR values are used to evaluate labeling performance.

\subsection{Manifold Curvature Approach} 
In this approach, similar to the SPM, the initial step involves constructing affordance label clusters for the feature vectors of the images. The learning phase is only grouping these vectors based on their respective affordance labels. When a new image is processed and a feature vector is generated using pre-trained feature extractor, manifold curvature calculation is carried out to perform affordance labeling.

Let $j$ denote the feature vector of a test image from $C_{validation}$, and let $C_i$ represent the previously established affordance label cluster for affordance $i$. Initially, $n$ vectors within $C_i$ with the smallest $l_2$ distances to vector $j$ are selected. These vectors are denoted as $p_{i,{1}}, p_{i,{2}}, \ldots, p_{i,{n}}$.

Once closest $n$ points from $C_i$ are found, matrix $\mathbf{\tilde{M}}_{i,j}$ is formed as follows:

$\mathbf{\tilde{M}}_{i,j} = \begin{bmatrix} p_{i,{1}} & p_{i,{2}} & \hdots & p_{i,{n}}\end{bmatrix}$

A second matrix i.e. $\mathbf{M}_{i,j}$ is formed as follows:

$\mathbf{M}_{i,j} = \begin{bmatrix} j \ \ \mathbf{\tilde{M}}_{i,j}  \end{bmatrix} =  \begin{bmatrix} j & p_{i,{1}} & p_{i,{2}} & \hdots & p_{i,{n}}\end{bmatrix}$






The objective here is, to understand the local effect of the new data point $j$ when it is introduced into a manifold. This is achieved  by comparing matrices $\mathbf{\tilde{M}}_{i,j}$ and $\mathbf{M}_{i,j}$. Following the construction of these matrices, we compute the skinny Singular Value Decomposition (SVD) for each of them, where the skinny SVD only uses the columns of $\mathbf{U}$ and rows of $\mathbf{V}^T$ that correspond to non-zero singular values. Equations~\eqref{eq:svd1} and~\eqref{eq:svd2} are based on skinny SVD: 
\begin{equation}
     \mathbf{U}_{i,j} \mathbf{\Sigma}_{i,j} \mathbf{V}^T_{i,j}=\mathbf{M}_{i,j}
\label{eq:svd1}
\end{equation}
\begin{equation}
    \mathbf{\tilde{U}}_{i,j} \mathbf{\tilde{\Sigma}}_{i,j} \mathbf{\tilde{V}}^T_{i,j}= \tilde{\mathbf{M}}_{i,j} 
\label{eq:svd2}
\end{equation}
After obtaining these decompositions, we can determine how the local subspace changes when the vector $j$ is included or excluded from the neighborhood. Since the columns of the $\mathbf{U}$ matrix represent unit basis vectors of the neighborhood, we can utilize the $\mathbf{U}$ matrices similar to what is done in finding the principal angles between two subspaces. Based on $\mathbf{U}_{i,j}$ and $\mathbf{\tilde{U}}_{i,j}$, matrix $\mathbf{R}_{i,j}$ is defined in equation~\eqref{eq:R}:
\begin{equation}
   \mathbf{R}_{i,j} = \mathbf{U}_{i,j}^T \mathbf{\tilde{U}}_{i,j} 
\label{eq:R}
\end{equation}
In this context, the diagonal elements of matrix $\mathbf{R}_{i,j}$ are $\cos\theta_{1,1}, \cos\theta_{2,2}, \ldots, \cos\theta_{n,n}$, where $\theta_{n,n}$ represents the angle between the $n^{\text{th}}$ columns of $\mathbf{U}_{i,j}$ and $\mathbf{\tilde{U}}_{i,j}$. The angle $\theta$ is calculated simply by summing the diagonal entries as shown in equation~\eqref{eq:thetadef} to indicate some sort of a cumulative agreement between 2 subspaces as shown in equation ~\eqref{eq:thetadef}:
\begin{equation}
    \theta = \arccos\sum_{k=0}^n r_{k,k}
    \label{eq:thetadef}
\end{equation}
However, note that the order of the columns in the $\mathbf{U}$ matrix is important, and their weight in calculating the original matrix is represented by $\mathbf{\Sigma}$. Therefore, equation~\eqref{eq:thetadef} can be refined to reflect the importance of agreement / disagrement between dominant (i.e. principal) directions to result in yet a more precise angle calculation~\cite{sekmen2022manifold}. Thus, $\mathbf{R}_{i,j}$ matrix is redefined as shown in equation~\eqref{eq:upR}:
\begin{equation}
   \mathbf{R}_{i,j} = (\mathbf{U}_{i,j} \mathbf{\Sigma}_{i,j})^T (\mathbf{\tilde{U}}_{i,j} \mathbf{\tilde{\Sigma}}_{i,j}) 
   \label{eq:upR}
\end{equation}
Then, $\mathbf{R}_{i,j}$ matrix is decomposed using SVD in equation~\eqref{eq:Rsvd}.
\begin{equation}
    \mathbf{U}_{i,j}^" \mathbf{\Sigma}_{i,j}^" \mathbf{V}_{i,j}^{"T}= \mathbf{R}_{i,j} 
    \label{eq:Rsvd}
\end{equation}
Lastly, an expression similar to equation~\eqref{eq:thetadef} is calculated as shown in the equation~\eqref{eq:thetares} for the updated $\mathbf{R}_{i,j}$ matrix:
\begin{equation}
    \theta_w = \arccos \frac{\sum_{k=0}^n\sigma_{{i,j}_{kk}}^"}{\sum_{k=0}^n\sigma_{{i,j}_{kk}}\tilde{\sigma}_{{i,j}_{kk}}}
    \label{eq:thetares}
\end{equation}
Where $\sigma_{{i,j}_{kk}}^"$, $\sigma_{{i,j}_{kk}}$ and $\tilde{\sigma}_{{i,j}_{kk}}$ are the $k^{\text{th}}$ diagonal entries of $\mathbf{\Sigma}_{i,j}^"$, $\mathbf{\Sigma}_{i,j}$ and $\mathbf{\tilde{\Sigma}}_{i,j}$ respectively. This angle-sum value serves as an indicator of the alteration within the local subspace with the introduction of $j$ to this local subspace. Here smaller angle values signify less variation or distortion in the local subspace, hence, it is an indication that $j$ is in harmony / agreement with that local subspace. In other words, if introducing the feature vector of a test image to the local neighborhood within an affordance set results in a minimal angle-sum value, it suggests that this new image can be affiliated with this affordance set. Therefore, if the angle-sum between the local subspaces of $\mathbf{M}_{i,j}$ and $\mathbf{\tilde{M}}_{i,j}$ is relatively small, it implies that the image represented by the feature vector $j$ should be labeled with affordance label $i$. This label corresponds to the group comprising the pre-defined vectors from the neighborhood from affordance cluster $i$, namely $p_{i,{1}}, p_{i,{2}}, \ldots, p_{i,{n}}$. Whole process of labeling via MCM is presneted in algorithm \ref{alg:mcm}.
\begin{algorithm}[H]
\caption{Affordance Labeling via MCM}\label{alg:mcm}
\textbf{Input:} $C_{learning}$, $C_{validation}$ :List of learning and validation feature vector sets,
\\
$n:$ number of neighbour vectors\\
$threshold$ : Threshold value for local subspace angle change
\textbf{Output:} $aff$: List of indices of objects labeled with corresponding affordance. 
\begin{algorithmic}
\State Initialize $aff$
\For{$C_{i,learning}$ in $C_{learning}$ with affordance group $i$}
    \State Initialize $aff_i$
    \For{$j$ in $C_{validation}$}
        \State $C_{neighbours} \gets n$ vectors closest to $j$ in $C_{i,learning}$
        \State $\tilde{\mathbf{M}}_{i,j} \gets horizontalStack(C_{neighbours})$
        \State ${\mathbf{M}}_{i,j} \gets horizontalStack(C_{neighbours}, j)$
        \State $\mathbf{U}_{i,j}, \mathbf{\Sigma}_{i,j}, \mathbf{V}^T_{i,j} \gets SVD(\mathbf{M}_{i,j})$
        \State $\mathbf{\tilde{U}}_{i,j}, \mathbf{\tilde{\Sigma}}_{i,j}, \mathbf{\tilde{V}}^T_{i,j} \gets SVD(\tilde{\mathbf{M}}_{i,j})$
        \State $\mathbf{R}_{i,j} \gets (\mathbf{U}_{i,j} \mathbf{\Sigma}_{i,j})^T (\mathbf{\tilde{U}}_{i,j} \mathbf{\tilde{\Sigma}}_{i,j}) $
        \State $\mathbf{U}_{i,j}, \mathbf{\Sigma}_{ij,R}, \mathbf{V}_{ij,R}^T \gets SVD(\mathbf{R}_{ij})$
        \State $diag\_sum_{i,j}^" \gets sum(diag(\mathbf{\Sigma}_{i,j}^"))$
        \State $diag\_sum_{i,j} \gets sum(diag(\mathbf{\Sigma}_{i,j}\mathbf{\tilde{\Sigma}}_{i,j}))$
        \State Calculate local subspace change angle as $\theta_{i,j} \gets \arccos(diag\_sum_{i,j}^"/diag\_sum_{i,j})$
        \If{$\theta_{i,j} \leq threshold$}
            \State Append $Arg(j)$ to $aff_i$
        \EndIf
    \EndFor
    \State Append $aff_i$ to $aff$
\EndFor
\State $\mathbf{return}$ $aff$
\end{algorithmic}
\end{algorithm}

\section{Results and Discussion}
In this section, the results of the proposed methods are tabulated and discussed. In this context, a selection of pre-trained neural networks, namely, ResNet-18, ResNet-50, ResNet-101, ResNet-152, ResNext-101, RegNetY, EfficientNetV2, ViT-L/16, and ViT-B/16 have been used in performance testing of the propsed methods. These  networks have been pre-trained on the ImageNet\cite{Imagenet} dataset. Due to the distinct architectures of these networks, their respective feature vector dimensions also differ.

For each of these pre-trained networks, the basis vectors, projection matrices, and threshold values for the affordance classes have been computed. Subsequently, utilizing subspace projection, the labels of the images have been determined, and an assessment of the labeling performance has been conducted. The results are given in Table \ref{net_comp}
\begin{table}[h]
\caption{\textbf{Affordance Labeling Performances of Proposed Methods on Selected Pre-trained Networks}}
\begin{center}
\begin{tabular}{|c|c|c|c|c|c|}
\hline
Network&Vector&\multicolumn{2}{c|}{SPM} &\multicolumn{2}{c|}{MCM}\\
\cline{3-6} 
 Name & Size& TPR(\%)&FPR(\%) &TPR(\%) &FPR(\%) \\
\hline
ResNet-18 & 512 &91.62 & 1.17& 96.16 & 0.61 \\
ResNet-50 & 2048 & 87.04 & 1.76 & 95.76 & 0.67\\
ResNet-101 & 2048 & 81.34 & 2.42 & 96.11 & 0.65\\
ResNet-152 & 2048 & 84.56 & 2.21 & 96.14 & 0.61 \\
ResNext-101 & 2048 & 93.23 & 0.76 & 95.55 & 0.72\\
RegNetY & 3024 &\textbf{94.07}  & 0.65 & \textbf{96.45} & 0.57\\
ViT-B/16 & 768&70.84  & 3.84 & 95.38 & 0.72 \\
ViT-L/16 & 1024& 74.60 & 3.37 & 95.54 & 0.73\\
EfficientNetV2 & 2560 & 90.95 & 1.00 & 96.35 & 0.57\\
\hline
\end{tabular}
\label{net_comp}
\end{center}
\end{table}

Our analysis reveals that even the benchmark method based Subspace Projection (SPM) which relies on the assumption that data coming from same labels are distributed on or around some linear subspaces resulted in performance that is comparable some of the results in the literature. It should be noted that variation from one network to another SPM performance significantly changes. On one hand, this limits the portability of this methods to different networks. On the other hand, higher performance using SPM can be considered as an indication of how that particular network has the ability to flatten input data coming from the same cluster as it progresses towards the end of the network. Therefore, in image based tasks where some subspace analysis tools are to be applied without additional training, RegNetY and ResNext-101 stand out as viable first alternatives.

Manifold Curvature Method (MCM) yielded results that are on par with the current state of the art in affordance labeling. Comparatively it is seen that MCM surpasses SPM by delivering consistently better results across various feature extractor networks. Notably, MCM also yields lower False Positive labeling rates. These findings suggest that while pre-trained networks trained on the ImageNet\cite{Imagenet} dataset can be used to assess subspaces of different classes decoded by one-hot encoding, dividing the outputs into multiple subspaces and evaluating for multi-hot encoded (a.k.a. multi label binarizer) cases introduces complexities due to subspace intersections. Additionally, SPM method is heavily reliant on the pre-trained network's capacity to group similar features into the same subspaces. In contrast, MCM excels because it evaluates data based on its appendability to existing data clusters, making it more transferable across different networks. Furthermore, by assessing the test vector's appendability to class clusters, this method offers flexibility for assigning multiple labels in intersection regions or no label at all. 

Khalifa and Shah\cite{khalifalarge} focused on the potential of the Large Scale Affordance Dataset utilized in this study by conducting experiments with various segmentation and labeling networks. Their efforts resulted in a mean accuracy of 63.38\% for the affordance segmentation task. Additionally, they explored transfer learning applications by adapting the last layer of pre-trained networks and training them to label affordances within the dataset they presented, achieving a mean accuracy of 91.83\%. Furthermore, they introduced a novel Visual Affordance Transformer\cite{shah} in a subsequent study, which led to updated affordance segmentation results. Intersection Over Union (IOU) was employed as a performance metric, with results provided for each affordance category. The highest IOU accuracy obtained was 85.65\% for the `grasp' affordance, while the lowest was 41.85\% for the `cut' affordance. Since this study focuses on labeling without segmentation, a comparison of labeling performances is feasible. SPM's labeling accuracy heavily relies on the pre-trained network utilized for feature extraction, with the highest True Positive Ratio reaching 94.07\% with RegNetY\cite{regnety}, surpassing Khalifa and Shah's\cite{khalifalarge} 91.83\% labeling accuracy. Moreover, MCM consistently yields better labeling results across various pre-trained neural networks, achieving a True Positive Ratio of 96.45\% with RegNetY\cite{regnety}, exceeding that of the SPM. Thus, MCM consistently outperforms the SPM in terms of True Positive and False Positive labeling outcomes. Table \ref{tbl:labels} showcases the performance of MCM and SPM in affordance labeling. While both approaches accurately predict some ground truth affordances (e.g., rollability and wrap-graspability of glue sticks), there are occasional errors (e.g., labeling illumination affordance for a ball). Interestingly, these methods also identify affordances not explicitly labeled in the ground truth but demonstrably applicable to the object. For instance, both methods correctly assign the `tip-push' affordance to the cell phone, even though the keyboard is not visible. This suggests the model recognizes a button-like element based on its location. Similarly, MCM assigns grasping and rolling affordances to the soda can, which are not present in the ground truth but are valid based on the object's physical properties. There are also instances of ambiguity, such as assigning the `open' affordance to a flashlight due to its cap-like part. While valid, MCM additionally captures the illumination affordance, highlighting its ability to identify multiple functionalities. The findings from Table 2 suggest that both MCM and SPM demonstrate promising performance in affordance labeling. Notably, their ability to capture affordances that are not explicitly labeled in the ground truth makes them valuable tools for exploring new potential interactions with objects. These methods can potentially reveal previously unconsidered affordances, leading to a broader planning space for robots as well as enhancing robot-object-human interaction. 
\begin{table*}
\caption{\textbf{Labeling Examples of Validation Set via SPM and MCM}}
  \centering
  \begin{tabular}{ | c | m{2.5cm} | m{2cm} | m{2.5cm}| m{2.5cm}|m{2.5cm}|}
    \hline
    Object Image & Ground Truth & Labeling Method & Correct Labeled Affordances Marked in Ground Truth &Correct Labeled Affordances not Present in Ground Truth & False Labeled Affordances\\ \hline
    \multirow{2}{*}{
    \begin{minipage}{.08\textwidth}
      \includegraphics[width=0.6\linewidth]{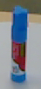}
    \end{minipage}
    }
    &
    \multirow{2}{*}{
    \begin{minipage}[t]{2.5cm}
      \begin{soloitemize}
        \item Grasp
        \item Open
      \end{soloitemize}
      \end{minipage}
    }
    & 
    SPM
    &
    \begin{soloitemize}
        \vspace{0.1cm}
        \item Grasp
        \item Open
        \vspace{0.1cm}
      \end{soloitemize}
    &
    \begin{soloitemize}
        \item Wrap-Grasp
      \end{soloitemize}
    &

    \\
    \cline{3-6}
    & & MCM 
    &
    \begin{soloitemize}
        \vspace{0.1cm}
        \item Grasp
        \item Open
        \vspace{0.1cm}
      \end{soloitemize}
    &
    \begin{soloitemize}
        \item Roll
      \end{soloitemize}
    &
    \\
    \hline
    \multirow{2}{*}{
    \begin{minipage}{.1\textwidth}
      \includegraphics[width=0.9\linewidth]{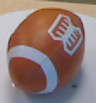}
    \end{minipage}
    }
    &
    \multirow{2}{*}{
    \begin{minipage}[t]{2.5cm}
      \begin{soloitemize}
        \item Wrap-Grasp
        \item Roll
      \end{soloitemize}
      \end{minipage}
    }
    & 
    SPM
    &
    \begin{soloitemize}
        \vspace{0.1cm}
        \item Wrap-Grasp
        \item Roll
        \vspace{0.1cm}
      \end{soloitemize}
    &
    
    &

    \\
    \cline{3-6}
    & & MCM 
    &
    \begin{soloitemize}
        \vspace{0.1cm}
        \item Wrap-Grasp
        \item Roll
        \vspace{0.1cm}
      \end{soloitemize}
    &
    \begin{soloitemize}
        \item Grasp
      \end{soloitemize}
    &
    \begin{soloitemize}
        \item Tip-Push
        \item Illuminate
      \end{soloitemize}
    \\
    \hline
    \multirow{2}{*}{
    \begin{minipage}{.1\textwidth}
      \includegraphics[width=0.86\linewidth]{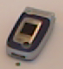}
    \end{minipage}
    }
    &
    \multirow{2}{*}{
    \begin{minipage}[t]{2.5cm}
      \begin{soloitemize}
        \item Grasp
        \item Display
      \end{soloitemize}
      \end{minipage}
    }
    & 
    SPM
    &
    \begin{soloitemize}
        \vspace{0.1cm}
        \item Grasp
        \item Display
        \vspace{0.1cm}
    \end{soloitemize}
    &
    \begin{soloitemize}
        \item Tip-Push
        \item Open
    \end{soloitemize}
    &

    \\
    \cline{3-6}
    & & MCM 
    &
    \begin{soloitemize}
        \vspace{0.1cm}
        \item Grasp
        \item Display
        \vspace{0.1cm}
    \end{soloitemize}
    &
    \begin{soloitemize}
        \item Tip-Push
      \end{soloitemize}
    &
    
    \\
    \hline
    \multirow{2}{*}{
    \begin{minipage}{.08\textwidth}
      \includegraphics[width=0.8\linewidth]{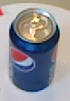}
    \end{minipage}
    }
    &
    \multirow{2}{*}{
    \begin{minipage}[t]{2.5cm}
      \begin{soloitemize}
        \item Wrap-Grasp
        \item Open
      \end{soloitemize}
      \end{minipage}
    }
    & 
    SPM
    &
    \begin{soloitemize}
        \vspace{0.1cm}
        \item Wrap-Grasp
        \item Open
        \vspace{0.1cm}
    \end{soloitemize}
    &
    
    &

    \\
    \cline{3-6}
    & & MCM 
    &
    \begin{soloitemize}
        \vspace{0.1cm}
        \item Wrap-Grasp
        \item Open
        \vspace{0.1cm}
    \end{soloitemize}
    &
    \begin{soloitemize}
        \item Grasp
        \item Roll
    \end{soloitemize}
    &
    \begin{soloitemize}
        \item Tip-Push
    \end{soloitemize}
    \\
    \hline
    \multirow{2}{*}{
    \begin{minipage}{.12\textwidth}
      \includegraphics[width=\linewidth]{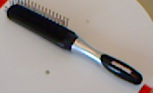}
    \end{minipage}
    }
    &
    \multirow{2}{*}{
    \begin{minipage}[t]{2.5cm}
      \begin{soloitemize}
        \item Grasp
      \end{soloitemize}
      \end{minipage}
    }
    & 
    SPM
    &
    \begin{soloitemize}
        \item Grasp
    \end{soloitemize}
    &
    
    &
    \begin{soloitemize}
        \vspace{0.1cm}
        \item Cut
        \item Open
        \item Grip
        \vspace{0.1cm}
    \end{soloitemize}
    \\
    \cline{3-6}
    & & MCM 
    &
    \begin{soloitemize}
        \vspace{0.1cm}
        \item Grasp
        \vspace{0.1cm}
    \end{soloitemize}
    &

    &
    
    \\
    \hline
    \multirow{2}{*}{
    \begin{minipage}{.125\textwidth}
      \includegraphics[width=\linewidth]{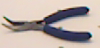}
    \end{minipage}
    }
    &
    \multirow{2}{*}{
    \begin{minipage}[t]{2.5cm}
      \begin{soloitemize}
        \item Grasp
        \item Grip
      \end{soloitemize}
      \end{minipage}
    }
    & 
    SPM
    &
    \begin{soloitemize}
        \vspace{0.1cm}
        \item Grasp
        \item Grip
        \vspace{0.1cm}
    \end{soloitemize}
    &
    \begin{soloitemize}
        \item Cut
    \end{soloitemize}
    &

    \\
    \cline{3-6}
    & & MCM 
    &
    \begin{soloitemize}
        \vspace{0.1cm}
        \item Grasp
        \item Grip
        \vspace{0.1cm}
    \end{soloitemize}
    &
    
    &
    
    \\
    \hline
    \multirow{2}{*}{
    \begin{minipage}{.125\textwidth}
      \includegraphics[width=\linewidth]{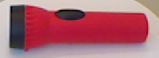}
    \end{minipage}
    }
    &
    \multirow{2}{*}{
    \begin{minipage}[t]{2.5cm}
      \begin{soloitemize}
        \item Grasp
        \item Tip-Push
        \item Roll
      \end{soloitemize}
      \end{minipage}
    }
    & 
    SPM
    &
    \begin{soloitemize}
        \vspace{0.1cm}
        \item Grasp
        \item Tip-Push
        \item Roll
        \vspace{0.1cm}
    \end{soloitemize}
    &
    \begin{soloitemize}
        \item Open
    \end{soloitemize}
    &
    \begin{soloitemize}
        \item Cut
    \end{soloitemize}

    \\
    \cline{3-6}
    & & MCM 
    &
    \begin{soloitemize}
        \vspace{0.1cm}
        \item Grasp
        \item Tip-Push
        \item Roll
        \vspace{0.1cm}
    \end{soloitemize}
    &
    \begin{soloitemize}
        \item Open
        \item Illuminate
      \end{soloitemize}
    &
    
    \\
    \hline
  \end{tabular}
  \label{tbl:labels}
\end{table*}

\section{Conclusion}
In this paper, we introduced two approaches for identifying affordance labels of object. These methods generate decisions by analysing feature vectors that are extracted from existing pre-trained networks using subspace projections and so-called manifold curvatures respectively. What is to be noted is that, neither of these proposed methods require any further training, which separates them from the ones existing in the literature. These methods have been evaluated using nine well-known pre-trained networks. Results in indicate the efficacy of both the Subspace Projection Method and the Manifold Curvature Method, achieving accuracies exceeding 94\% and 95\% respectively. This demonstrates the practical applicability of these approaches in affordance labeling. Furthermore, our observations suggest that the proposed methods possess the capability to discovering affordances of objects not explicitly labeled in the ground truth information. This is particularly evident in instances such as the identification of the `roll' affordance for cylindrical objects like cans, which are typically not labeled as `roll' in the ground truth dataset. Conversely, the detection of `cut' affordance for pliers may not be as readily apparent. Such findings show the potential of our methods for affordance exploration. Additionally, owing to their reliance on feature vectors and vector operations, these methods offer flexibility in accommodating new affordance categories and incorporating new data into existing affordance groups easily, i.e. without the need of any training. 
\clearpage
\bibliographystyle{ieeetr} 
\bibliography{afford_arxiv}
\end{document}